# Bengali Common Voice Speech Dataset for Automatic Speech Recognition

*Samiul Alam*[1*,2], *Asif Sushmit*[1*,3], *Zaowad Abdullah*[1], *Shahrin Nakkhatra*[1], *MD. Nazmuddoha Ansary*[1], *Syed Mobassir Hossen*[1], *Sazia Morshed Mehnaz*[1], *Tahsin Reasat*[1,4], *Ahmed Imtiaz Humayun*[1,5]

[1]Bengali.AI [2]Michigan State University [3]RPI [4]Vanderbilt University [5]Rice University

*equal contribution

## Abstract

Bengali is one of the most spoken languages in the world with over 300 million speakers globally. Despite its popularity, research into the development of Bengali speech recognition systems is hindered due to the lack of diverse open-source datasets. As a way forward, we have crowdsourced the *Bengali Common Voice Speech Dataset*, which is a sentence-level automatic speech recognition corpus. Collected on the *Mozilla Common Voice platform*, the dataset is part of an ongoing campaign that has led to the collection of over 400 hours of data in 2 months and is growing rapidly. Our analysis shows that this dataset has more speaker, phoneme, and environmental diversity compared to the OpenSLR Bengali ASR dataset, the largest existing open-source speech dataset. We present insights obtained from the dataset and discuss key linguistic challenges that need to be addressed in future versions. Additionally, we report the current performance of a few Automatic Speech Recognition (ASR) algorithms and set a benchmark for future research.

## 1. Introduction and Related Work

With the growing popularity of virtual platforms for communication and learning, the field of speech analysis has seen widespread development. Apart from large contributions in technology accessibility through voice assistants (1), Text-to-Speech (TTS) and Automatic Speech Recognition (ASR) (also known as Speech-To-Text (STT)) have ubiquitous applications across the spectrum, e.g., Automatic Speech Assessment for language education (2), language disorder assessment and therapy (3), and assistive agriculture (4). Despite extensive theoretical and computational studies done on modeling Bengali phonology and the advent of powerful deep learning networks, language technologies like the ones mentioned above have not been realized for Bengali due to the lack of datasets.

Large deep learning methods require large-scale annotated datasets, a requirement that is made even more imperative by the diverse accents and unique morphology of the Bengali language. Over the years, various datasets have been assembled each having its own limitations regarding speaker diversity, size and public availability (5; 6; 7; 8). To facilitate research in this direction, the Bengali Common Voice Speech Dataset: a collection of over 400 hours of transcribed audio recordings of spoken Bengali sentences was gathered from all over Bangladesh and India via community outreach and collaborative efforts. The corpus is part of an ongoing campaign to collect audio recordings by Bengali.AI with a target of reaching 5000 hours of collected audio by 2023. In this paper, we discuss some important linguistic challenges for developing Bengali speech recognition systems and illustrate key insights observed from the data collected. We also tabulate the benchmark Word error rate (*WER*) of an HMM-GMM based ASR algorithm. This paper will be collaboratively updated with newer benchmarks and for newer releases of the Common Voice Speech Corpus

The paper is organized as follows: Section 2 explains the key features of Bengali speech and the challenges they pose in the design of ASR and TTS systems. Section 3 describes the dataset in detail and illustrates important statistics showcasing the diversity of the data. Section 4 further analyses the data by comparing its feature diversity with Google's OpenSLR dataset. Section 5 shows preliminary benchmarks on the dataset for baseline models. Finally, Section 6 presents our conclusions. For the ease of non-bengali readers, we have supplied the IPA transliterations (9) of Bengali words.

## 2. Linguistic Challenges in ASR and TTS modeling of Bengali

Bengali is a language with high linguistic complexity due to its writing system, inflectional morphology, gemination, and a high number of diphthongs and triphthongs (10). It is crucial to know these concepts to effectively design and deploy speech recognition algorithms for the Bengali language. In this section, we discuss some of the unique linguistic challenges that Bengali poses, compared to Romance or Germanic languages.

### 2.1. Grapheme Diversity

One of the most unique features of the Bengali writing system is the use of orthographic syllables as graphemes. All alpha-syllabary or *Abugida* languages follow this trait. We therefore interchangeably use the terms *orthographic syllables* and *graphemes* throughout this text. For Bengali, the number of unique orthographic syllables and hence, the number of possible words are very high. Bengali contains over 1300 commonly used orthographic syllables whereas $\sim$ 15 million graphemes are theoretically possible (11). Many of these occur in rare words and also due to spelling mistakes(12). The Bengali Unicode system also allows duplicate ways to encode the same alphabet, which requires normalization (13) and also adds to the complexities surrounding grapheme diversity in text.

In Bengali, there are 168 commonly used grapheme roots, and of them, 80 are consonant conjuncts consisting of 2, 3 or higher consonants. A high number of consonant conjuncts leads to challenges in grapheme to phoneme mapping which depreciates the performance of TTS models, as the same conjuncts often have different pronunciations based on the context. For example, সুস্মিত is pronounced ʃuʃmit̪ whereas বিস্মিত is pronounced biʃʃit̪o, despite having the same grapheme স্মি in the middle.

## 2.2. Inflection

In linguistics, *inflection* is the process of forming word variants using a template word based on grammar (e.g., variants based on tense) or sentence semantics (e.g., variants based on the gender of nouns) (14). Like most other Indic languages, Bengali is highly inflectional. Nouns have over 100 case markers/inflection variations and verbs are found to have more than 40 unique variations. This makes the number of whitespace-tokenized 'words' occur very frequently, on the level of tens of millions compared to 188k words in English (15). Bengali morphology also allows compound word creation which leads to a high number of Out of Vocabulary (OOV) words. This issue causes challenges in TTS and ASR modeling, especially for low-resource languages (16) like Bengali.

## 2.3. Aspiration

*Aspirated* consonants খ, ঘ, ছ, ঝ, ঠ, ঢ, থ, ধ, ফ, ভ) [kʰ, gʱ, cʰ, ɟʱ, tʰ, dʱ, t̪ʰ, d̪ʱ, pʰ, bʱ] are consonants followed by voiceless glottal fricatives (17). Aspirated consonants pose challenges both in ASR and TTS modeling due to their subtlety in pronunciation. This, coupled with exaggerated aspiration in breathy voices, makes modeling real-life speech extremely hard (18). Moreover, people with strong accents as well as non-native speakers, pronounce these aspirated sounds differently and hence make generalization difficult. (19).

## 2.4. Gemination

*Gemination* is an instance of uttering a consonant for a longer period of time due to its consecutive use in a sentence (20). For example, বাদ দাও[bad̪ d̪a͡o̯) may also be pronounced বাদ্দাও[bad̪d̪a͡o̯), where the latter is geminated; the difference between these two pronunciations are subtle in natural speech but are quite different in text. In Bengali, gemination may occur both in-between words and within a single word. But it is more common between two consecutive words which start and end with similar phonemes. While this linguistic feature poses challenges in determining word boundaries for ASR algorithms for many languages (21), the modeling challenges posed for Bengali are not studied in detail in the literature. Roughly, gemination in Bengali can be summarized into the following bins:

1. Assimilation: Two sounds in a word can influence each other and reach a level of equality. জন্ম>জম্ম[ɟɔnmo > ɟɔmmo], কাঁদনা>কান্না[kãd̪na > kanna]
2. Progressive Assimilation: Here the previous sound of the word influences the later sound of the word. পদ্ম>পদ্দ[pɔd̪d̪o > pɔd̪d̪o], লগ্ন>লগ্গ[lɔgno > lɔggo]
3. Long Consonant: Sometimes to emphasize something the sounds are uttered dually in a word. পাকা>পাক্কা[paka > pakka], সকাল>সক্কাল[ʃɔkal > ʃɔkkal]
4. Deletion of Consonant Diacritic R (র): Due to the deletion of র-কার the later sound of the previous র is often uttered as a double sound of that consonant. তর্ক>তক্ক[t̪ɔrko > t̪ɔkko], মারল>মাল্ল[marlo > mallo],করলাম>কল্লাম[korlam > kollam]
5. Juncture: Juncture in Bengali follows some specific rules and forms gemination, such as: বন + ওষধি = বনৌষধি [bɔn + oʃod̪ʱi = bɔno͡u̯ʃod̪ʱi] মরু + উদ্যান = মরূদ্যান[moru + ud̪d̪an = morud̪d̪an] বিপদ+জাল = বিপজ্জাল [bipɔd̪ + ɟal = bipɔjjal]

## 2.5. Diphthong and Triphthong

A *diphthong* is a combination formed with two vowels in a single syllable. The semi-vowel of the diphthong is placed either on the onset or the coda of the syllable. Thus diphthong is the linguistic summation of vowels plus glide. The exact number of Bengali diphthongs has varied from expert to expert (22). According to Sukumar Sen (23), there are only two Bengali diphthongs with characters assigned to them (ঐ, ঔ). Suniti Kumar Chatterjee (24) had however pointed out 25 diphthongs while Muhammad Abdul Hai had opined that the Bengali language can have up to 31 diphthongs among which 19 are regular and 12 are irregular (25). Bengali has a far higher number of diphthongs than English (which has 10 diphthongs).

Similarly, we have triphthongs. "Triphthong is a combination of three vowel sounds where the first vowel glides to the second which again glides to the third." as defined by Barman et al. (26). In English, there are 5 Tripthongs whereas Bengali has 17.

Apart from these, there are many grammatical features in Bengali that make ASR and TTS modeling hard for Bengali. Brahmic Schwa deletion ambiguity (27) i.e. many Bengali words dropping their trailing vowel sounds are hard to model without a proper language model. The high number of homographs in Bengali (28) (29) are also related to this issue making modeling more challenging. Nasality (30) is often skipped while pronouncing by native Bengali speakers, but their written forms often explicitly include the markers of it. Also, it has been reported (31) that the phonological phrases in Bengali correspond to no plausible constituent of syntactic representation. This creates problems in modeling speech ensuring proper prosody. In Bengali speech, the phone(s) associated with glide (semivowel, glide, or semi-consonant is a sound that is phonetically similar to a vowel sound but functions as the syllable boundary), diphthongs (sound formed by the combination of two vowels) and emphasizers (modifier that serves to enhance and give additional emotional context) often sound similar. As these have different written representations, distinguishing them often requires high contextual information and this also poses challenges while modeling (32).

One unavoidable instance in Bangla sentence structure is the changing position of morpheme in the sentence. Bangla Language follows the SOV pattern of sentence structure. But in literature or daily life discourse, this pattern often alters to SVO, VSO, and OVS. (33) (34)

Also, there are Bengali speakers who use a high number of foreign/loan words from different languages and the pronunciation and spelling of these are often not standardized. This also causes challenges while working with real-life/diversified data (35)

## 3. Bengali Common Voice Corpus v9.0

The corpus has 231,120 *samples* from 19,817 *contributors* resulting in 399 *hours* (56 hours or 14% validated by one or more users) of speech recordings. Each audio clip is accompanied with a *'sentence'* annotation and additional metadata, namely: *'up votes', 'down votes', 'age', 'gender',* and *'accents'*. The sentence annotation is present for all the audio clips whereas, rest of the metadata comes from only the data contributors that have opted to provide it by logging in to the Mozilla common voice platform (36).

### 3.1. Collection Protocol

The Mozilla common voice platform allows data contributors to record single-channel audio data at 48kHz by reading prompts. The source sentences for the prompts were (1) randomly crawled from Wikipedia (37) using the Mozilla sentence collector or (2) provided to the sentence collector manually. All the text prompts are collected and distributed under a CC0 license. There is a fixed time constraint for reading the prompts. If sentences are too long, they contain obscene language, or if they are incorrect, the contributor has the option to skip or report a prompt for review. The collected recordings are released under a CC0 license after minimal post-processing.

### 3.2. Validation of Data

Through the validation platform of Mozilla common voice, 56 hours equivalent of data has been independently evaluated by community contributors that amount to 14% of the total data. The testing and validation splits both contain data from this validated portion of the dataset. Separate metadata with validated-invalidated flags is provided by Mozilla along with the release.

To verify recording quality, we survey 720 recordings randomly from the whole dataset (including samples from both the validated and invalidated portions) for qualitative analysis with a *4.8%* margin of error with *99%* confidence. We see that of the 720 recordings, 0.417% are unfinished recordings, 0.833% recordings are too fast to comprehend, 0.417% had loud background noise, 4.444% recordings were muffled, 0.417% were incomprehensible, 1.25% recordings had stuttering, and 0.694% contained no speech data. We share the details with clip names in the GitHub repository associated with this work.

### 3.3. Corpus Statistics

The text corpus contains 135,625 unique sentences. Each sentence has 7.12 words on average and each word contains 3.24 graphemes and 4.95 Unicode characters on average.

Table 1: *Recording time mean and variance of samples from the training, validation and test split. The TrainBAI is the accumulated training split containing the evaluated and non evaluated samples.*

| Split | Mean (s) | Variance (s) |
|---|---|---|
| TrainBAI | 6.20 | 3.66 |
| Validation | 6.28 | 3.32 |
| Test | 6.46 | 3.48 |

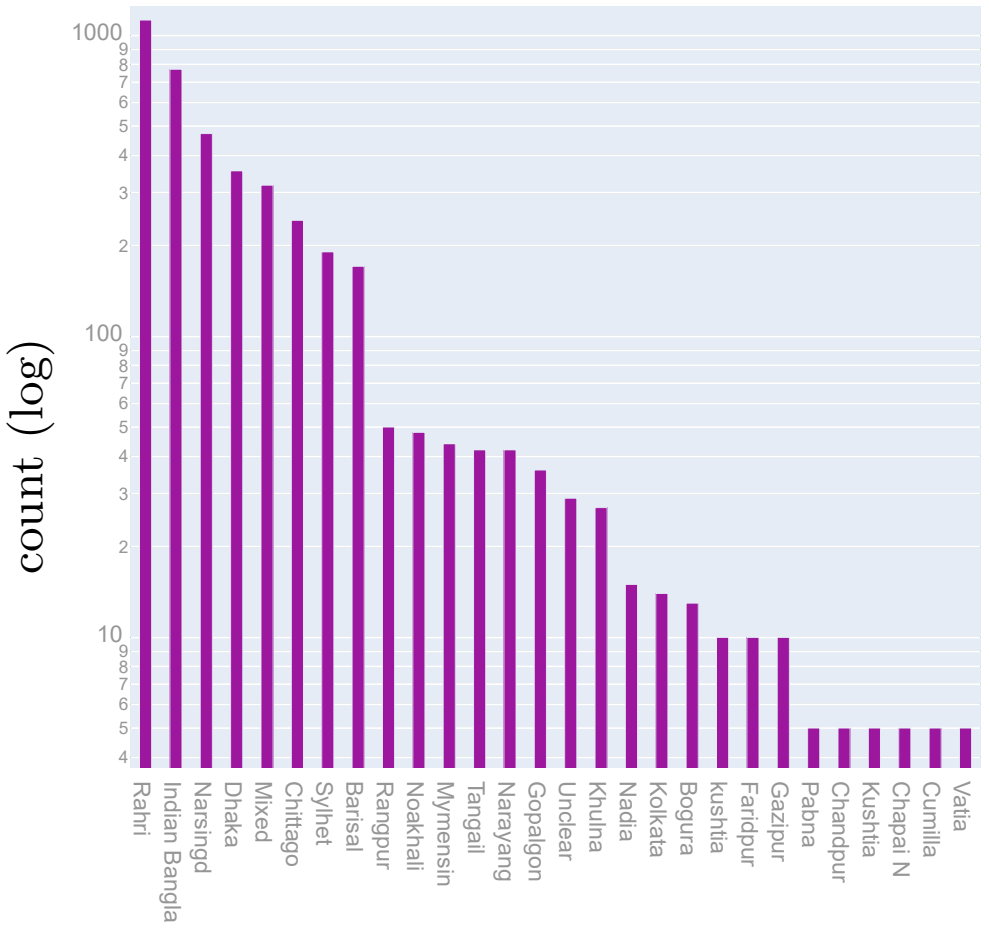

Figure 1: *Sample count of different self reported accents for 38.8% samples of the Common Voice speech corpus.*

### 3.4. Word and Grapheme Diversity

The corpus contains 47,932 unique words and 1,490 unique graphemes. In this version of the dataset, the sentences are mostly acquired from Wikipedia. Therefore, topic diversity in the text is high. However, interrogative sentences, exclamatory sentences, lyrical structures and patterns, and rhymes which are widely present in the Bengali language (38), are not as commonly present in the dataset.

### 3.5. Speech Diversity

The average number of recordings per sentence is 1 for the dev and test set but slightly higher at 1.58 for the train set. The mean and variance of the sample recording times are similar for the three splits (Table 1). The training data contains approximately 84.3% male and 15.7% female samples.

### 3.6. Accent Statistics

Among the data contributors who have self-reported their accents, we find out that 38.8% of the data represent 'standard' accents of Bengali. After cleaning and mapping the accent meta-data, there are 28 different accent types including Bengali accents from different regions of both Bangladesh and India. Not all the volunteers provided their accent data, so the reported data only reflects a subset of the whole speech corpus. The stats on the accent of the corpus is shown in Fig. 1.

# 4. Comparison with Available Datasets

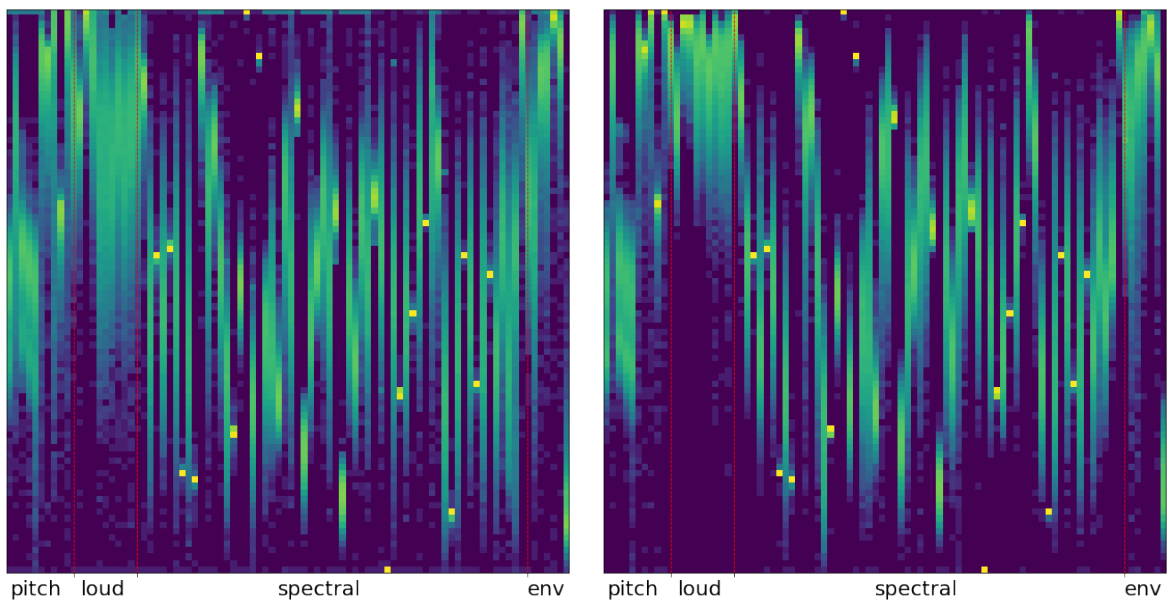

Figure 2: *Geneva feature (39) log-histograms for 10,000 samples from the Common Voice Bengali Dataset* ***(left)*** *and the OpenSLR Bengali ASR Dataset* ***(right)****. The feature set consists of pitch, loudness, spectral and envelope features separated via red markers in the figures. Common Voice has long-tailed distributions compared to OpenSLR due to higher speaker and environment diversity.*

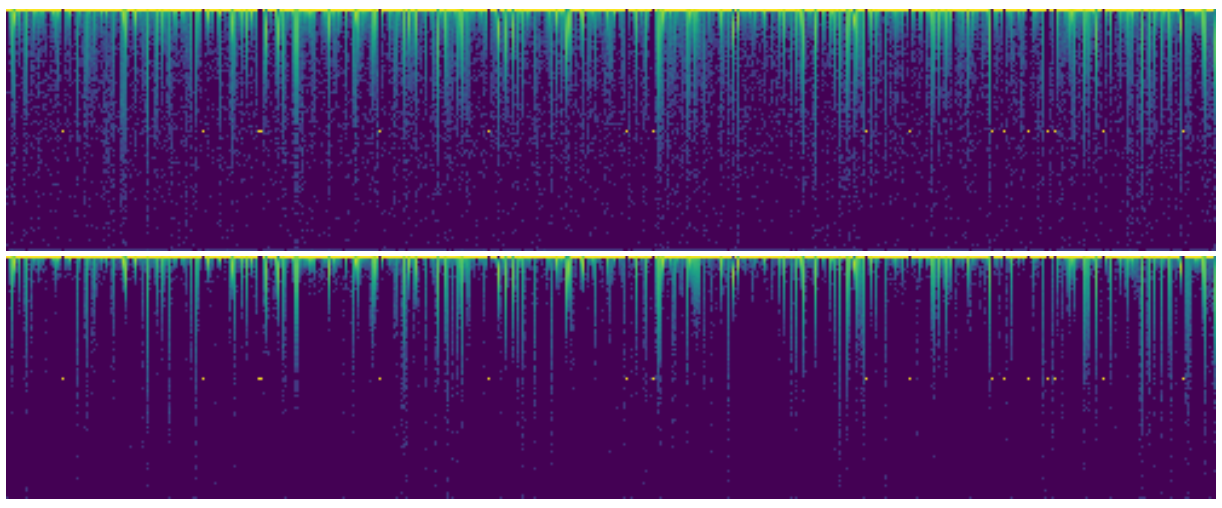

Figure 3: *Log-histograms of word recognition features for 10,000 samples from the Common Voice Bengali Dataset* ***(top)*** *and the OpenSLR Bengali ASR Dataset* ***(bottom)****. Features are extracted using a SpeechVGG (40) pre-trained on 1000 hours of English speech. The top histograms portray higher feature variability which is indicative of greater phoneme diversity in the Common Voice dataset.*

To date, there are 5 datasets of continuous sentence/utterance level speech corpus of Bengali language that can be found in the literature that contains more than 100 hours of speech data. Of those only 2 are publicly available. Also, the dataset mentioned in (41) is synthetically prepared. In Table 2 comparative information is shown on the prominent speech corpus and information on the newly released 399 hour Common Voice Bengali speech corpus, which is the *largest* and most *diversified publicly available* speech corpus for Bengali.

To assess the feature diversity of the speech in the Common Voice dataset, we extract Geneva speech features (39) and SpeechVGG word recognition features (40) for 10,000 samples of the Common Voice dataset and compare it with the OpenSLR speech dataset. In Fig. 2 and Fig. 3, we present visualizations that show higher diversity of features in the Common Voice dataset. Here we discuss some of the feature differences which we find insightful from our analysis. We also provide histograms for the features in Fig. 4.

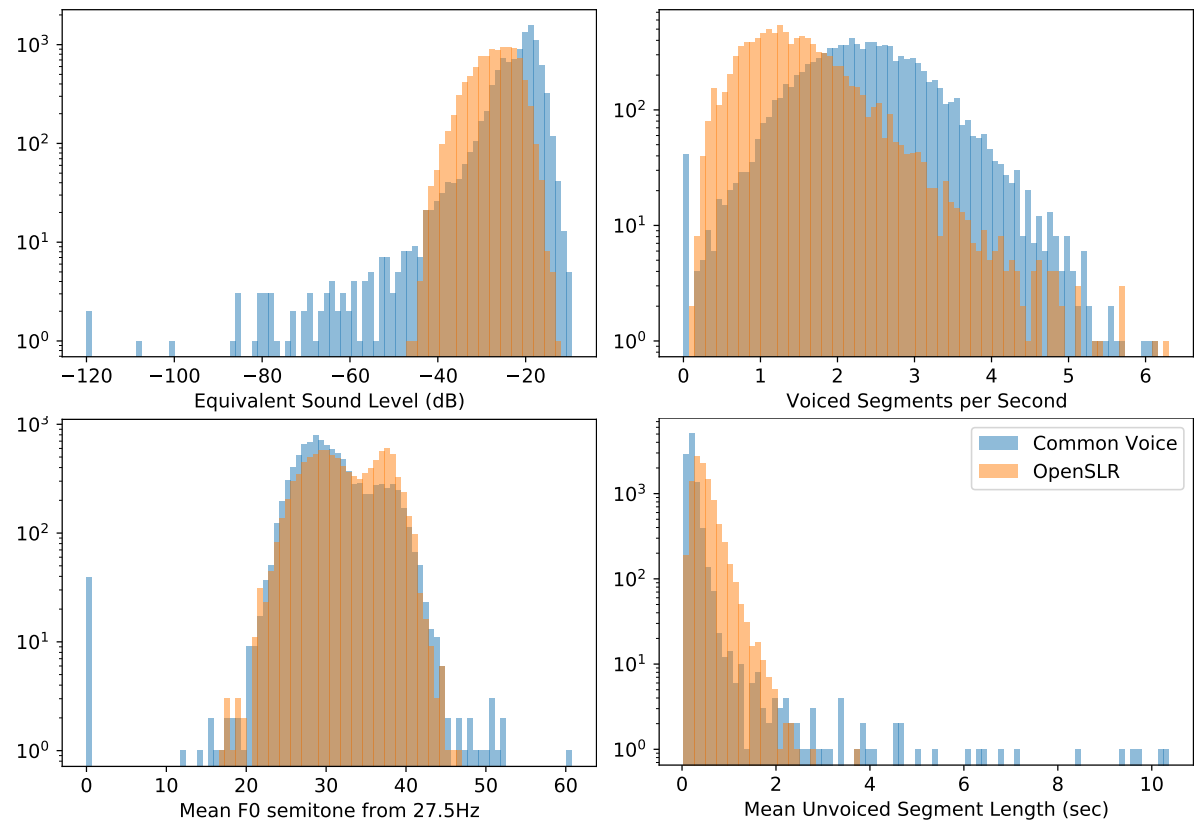

Figure 4: *Feature Histogram comparisons between Common Voice and OpenSLR datasets. Common Voice has varying sound levels due to uncontrolled recording environments compared to OpenSLR. It has higher average voice segments per second possibly due to the fixed recording time constraint on the collection portal.*

**Voiced Segments Per Second.** The minimum number of voiced segments per second in OpenSLR samples is 6.31 with a median of 1.36. Whereas, for Common Voice, the minimum is 0 and the median is 2.35. In Fig. 4-top-right, we can see a considerable number of samples with 0 voiced segments per second. This shows that Common Voice could have silent recordings or recordings with no vocalization, indicating a requirement for further curation. The median however for Common Voice is higher than OpenSLR, indicating that the sentences used in Common Voice require a higher frequency of vocalization compared to the phrases used in OpenSLR.
**Equivalent sound level DB.** We see that the sound level for OpenSLR has a lower dynamic range than Common Voice; with an upper and lower limit of -12 DB to -46 DB for OpenSLR and -9 DB to -120 DB for Common Voice. This is due to the fact that OpenSLR was acquired in a studio setting whereas Common Voice is crowdsourced from uncontrolled environments.
**Mean Unvoiced Segment Length.** The mean unvoiced segment length for non-empty recordings, quantifies the pauses in vocalization while recording. While Common Voice samples have smaller pauses compared to OpenSLR.
**Mean F0 semitone.** The Mean F0 semitone from 27.5Hz is the perceived pitch of recordings and varies based on gender and age. In Fig 4., we can see that it is bimodal for the OpenSLR dataset but for Common Voice, one of the modes are larger than the other. The reason for this could possibly be the higher gender imbalance in Common Voice compared to OpenSLR, with a larger fraction of masculine (lower pitch) recordings.

# 5. Benchmarking

We benchmark the dataset for ASR using Kaaldi (HMM-GMM) (43). We use the train-val-test splits provided in the original form of the Common Voice dataset, but include the unvalidated data into the training pipeline and also de-duplicate the sentences from the splits. The models will be progressively retrained, fine-tuned, and released in the aforementioned Github repo.

Table 2: *Comparison of Common Voice-Bengali Speech Corpus with other existing speech corpus. Detail on the other datasets taken from (42)*

| Year | Corpus Name | Size of Dataset | No. of Speakers | Availability |
|---|---|---|---|---|
| 2012 | IARPA-babel103b-v0.4b (5) | 215 hours | Not known | Not Publicly Available. Access per application. |
| 2014 | LDC-IL (6) | 138 hours | 240 males, 236 females | Not Publicly Available. |
| 2018 | OpenSLR (7) | 229 hours | 323 males, 182 females | Publicly Available under Attribution-ShareAlike 3.0 Unported (CC BY-SA 3.0 US) |
| 2020 | Bengali Speech Corpus from Publicly Available Audio & Text (41) | 960 hours | 268 males, 251 females | Not Publicly Available |
| 2020 | Subak.ko (8) | 241 hours | 33 males, 28 females | Not Publicly Available. |
| 2022 | **Common Voice Bengali Corpus V9.0** | **399 hours** | **20k Speakers** | **Publicly Available Under CC-0 Public Domain Attribution.** |

### 5.1. Dataset Split

The validated part of this dataset (56 hours) has been split into roughly 80% training, 10% testing, and 10%validation data. This is the train-test-eval split from the Mozilla Common Voice portal. While performing the split, it has been ensured that one speaker's recording is only in one split to make a fair evaluation of speaker generalization(36). While studying the dataset we propose the Bengali.AI Split for experimenting with the full (399 hours) dataset, which includes both the human validated and unvalidated data. For this, we have included all the non-validated data in the training set in our split while ensuring that no sentences that are present in the test set are present in the training set.

### 5.2. Result

For benchmarking the results we are using Word Error Rate (WER) and Character Error Rate (CER) as metrics. We are using Kaldi (HMM-GMM) model with standard hyper-parameters without any aggressive optimization techniques. We obtain 24.4 WER for this model. We also evaluate the performance of a benchmark model (Wev2vec2) trained on the OpenSLR dataset (44). On the test split of the Mozilla Common Voice Bengali dataset, the model shows 39.291 WER and 13.856 CER. We will progressively fine-tune these models and update them in the next versions of this paper. We release the training and evaluation codes and the custom splits here: https://github.com/BengaliAI/commonvoice-bangla.

## 6. Conclusion

In this paper, we reported the linguistic challenges of Bengali speech modeling and presented our analysis of the feature diversity and modeling on the 399 hour Bengali Common Voice Speech dataset. We show that this is the largest publicly available sentence-level ASR dataset, with greater feature diversities compared to its contemporaries. This dataset is promising for federated learning and many types of experimental modeling which are needed to be explored in the future. The dataset is under development, so eradicating the currently existing issues such as gender bias, corpus sentence structure diversification, etc will be needed in the next version of the dataset.

## 7. Acknowledgements

We thank the Mozilla foundation and their Common Voice project for the data collection framework.